\documentclass[pdflatex,sn-mathphys-num]{sn-jnl}

\usepackage{graphicx}%
\usepackage{multirow}%
\usepackage{amsmath,amssymb,amsfonts}%
\usepackage{amsthm}%
\usepackage{mathrsfs}%
\usepackage[title]{appendix}%
\usepackage{xcolor}%
\usepackage{textcomp}%
\usepackage{manyfoot}%
\usepackage{booktabs}%
\usepackage{algorithm}%
\usepackage{algorithmicx}%
\usepackage{algpseudocode}%
\usepackage{listings}%
\usepackage{bm}
\begin{document}
\title{Learning Beyond Rigid Manipulation with Hybrid Soft–Rigid Robots}
\author{
\centering
Mariano Ram\'irez Montero$^{1,\ast}$, Ebrahim Shahabi$^{1}$, Giovanni Franzese$^{2}$\\
Jens Kober$^{3}$, Barbara Mazzolai$^{4}$, Cosimo Della Santina$^{1}$\\[2ex]
{\normalsize
$^{1}$Cognitive Robotics, Delft University of Technology, Delft, The Netherlands\\
$^{2}$Technology Innovation Institute (TII), Abu Dhabi, United Arab Emirates\\
$^{3}$Institute for Artificial Intelligence, University of Stuttgart, Stuttgart, Germany\\
$^{4}$Bioinspired Soft Robotics Lab, Istituto Italiano di Tecnologia, Genova, Italy\\[1ex]
$^\ast$To whom correspondence should be addressed;\\
E-mail: M.RamirezMontero-1@tudelft.nl
}
}
\abstract{Rigid–soft robot hybrids promise to unite the precision and repeatability of rigid manipulators with the compliance and adaptability of soft structures. Yet, full-fledged rigid–soft manipulators remain largely unexplored, in part because the intelligence required to exploit this combination is still missing. Here, we introduce a hybrid robotic system obtained by the series interconnection of a rigid arm and a soft manipulator, and systematically examine its ability to perform both established and previously inaccessible manipulation tasks. We show that a tailored imitation learning algorithm—generally applicable to this class of hybrid systems—leverages the complementary roles of rigid precision and soft embodiment to acquire manipulation skills and generalize them in a single shot across task variations. As a result, the system matches or exceeds the performance of rigid robots in classical tasks such as stacking, while enabling qualitatively new behaviors, including grasping objects through narrow openings.}
\maketitle
\begin{figure}[th!]
    \centering
    \includegraphics[width=0.98\textwidth]{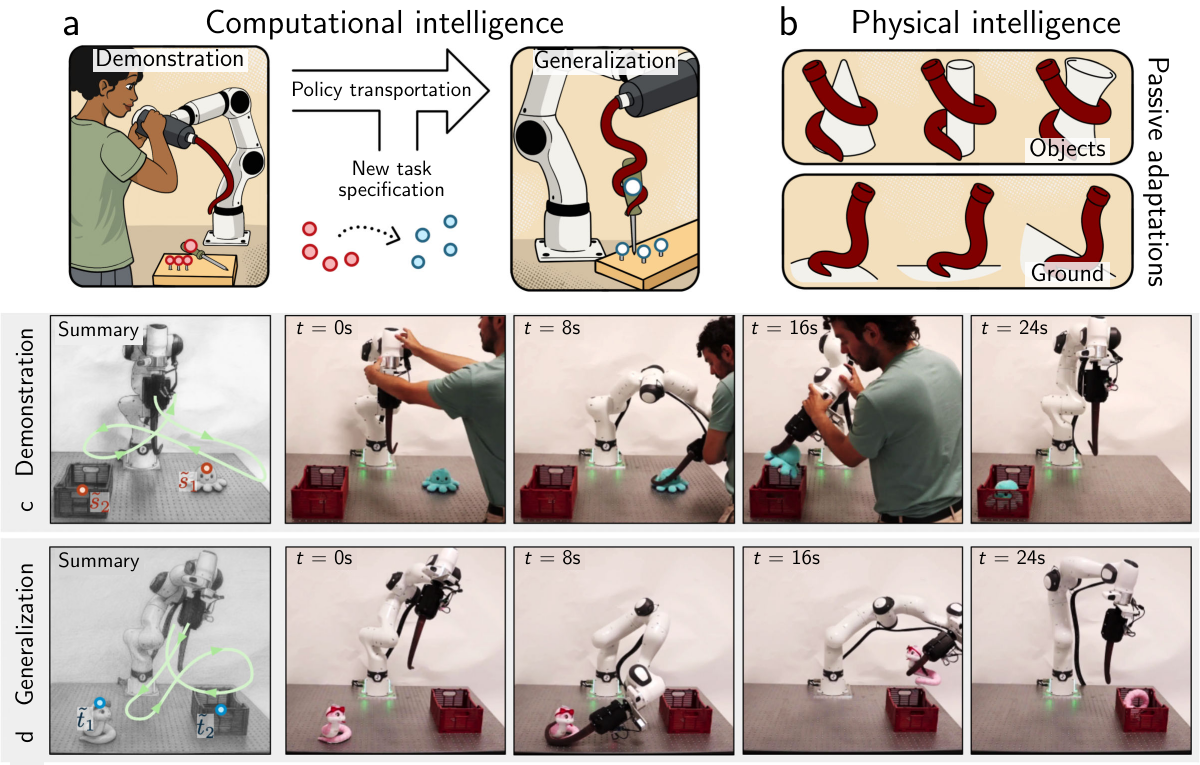}
    \caption{\textbf{Overview of the proposed hybrid system combining computational and physical intelligence.} \textbf{a} A human provides a single kinesthetic demonstration by moving the rigid part of the proposed rigid-soft hybrid manipulator. The learned policy is then transported to a new task instance using keypoint-based deformation, enabling generalization without retraining. The errors introduced by the transportation process, together with possible variations in the physical surroundings (e.g., different screwdriver), are absorbed by the soft arms' physical intelligence. \textbf{b} The soft arm can adapt passively to interactions with objects and the ground, compensating for uncertainties and variations in the environment and in the relative positioning of the soft arm base without requiring changes to the control input. Panels \textbf{c} and \textbf{d} show a side-by-side comparison of demonstration and generalization for a pick-and-place task. The first column reports a summary of the two keypoints (source $\tilde{s}_i$ and target $\tilde{t}_i$) used in this task, and of the two trajectories (demonstrated in \textbf{c} and transported in \textbf{d} at the rigid robot end effector.}
    \label{fig:cover_pic}
\end{figure}
Inspired by nature, soft robots exhibit embodied intelligence that enables them to adapt and even make decisions through purely physical processes \cite{rus2015design, laschi2016soft, della2020soft, xie2023octopus, comoretto2025physical, milana2025physical}. This capability has spurred progress in fields like medicine \cite{wang2024clinical}, environmental monitoring \cite{del2024growing}, and underwater exploration \cite{li2023bioinspired, del2026peripheral}.
Yet, despite their remarkable abilities and considerable efforts in modelling and control \cite{huang2020dynamic,mengaldo2022concise,della2023model,haggerty2023control}, soft robots struggle with reliable, precise, and repeatable motion. Rigid robots, by contrast, excel in high-precision, high-repetition tasks typical of manufacturing and assembly \cite{chen2024machine,ma2025learning,kirschner2025categorizing}.

We argue that, rather than attempting to make soft robots more precise \cite{della2020model} or rigid robots more naturally compliant \cite{ficuciello2015variable}, integrating the strengths of these two domains results in systems that are both mechanically intelligent and reliable.
To some extent, this insight is already well recognized.
Indeed, soft grippers integrated into rigid manipulators \cite{homberg2019robust, wang2020topology, dilibal2021additively, wang2023bioinspired,yue2025embodying, liu2023hybrid} — possibly combined with moving bases \cite{stokes2014hybrid, appius2022raptor, guo2024powerful} — have become the most successful implementation of soft robotic technology to date, with validations in real-world applications.
At the same time, more complex hybrids are surprisingly uncommon. If we exclude the use of small-scale soft robots as surgical tools \cite{cianchetti2018biomedical}, there are virtually no examples in the literature of full-fledged integration of rigid and soft manipulators.
This has been changing in the past year. After initial explorations in simulation \cite{mathew2023sorosim, morasso2025computational}, physical realizations of such hybrid platforms have only recently begun to emerge \cite{huang2025grasping, wang2025spirobs, koe2025learning, peng2025dexterous}.
However, these efforts center on relatively stiff systems in which the continuum part consists of rigid components or interacts with the environment solely at the tip.
In contrast, we combine a rigid robotic manipulator with a continuous, fully soft body containing embedded tendons, and we perform skills that involve full-body grasping and high contact.

Still, realizing the full potential of such hybrids requires more than mechanical integration—it requires intelligence.
Prior work has only begun to explore this direction; \cite{koe2025learning} proposes a non-quasistatic pose controller for a rigid 2-DoF arm with a continuum attachment, \cite{peng2025dexterous} addresses whole-body planning for a drone with a segmented soft extension, \cite{wang2025spirobs} investigates low-level control of spiral-shaped soft robots, and finally \cite{huang2025grasping} focuses on reproducing certain bio-inspired grasping strategies and on the design of the continuum robot.
An important distinction is that these approaches focus on lower-level intelligence: how to best design and control the soft part to achieve grasping.
While this lower-level is an important aspect of hybrids, we instead focus on mid-level intelligence, at a higher level of abstraction.
Instead of focusing only on grasping, we autonomously complete tasks such as stacking or pick-and-place through narrow gaps in varying configurations.

Moreover, instead of designing specific grasping behaviours, we make our low-level control task easier by exploiting imitation learning, a technique that has remained under-explored in the context of soft robotics \cite{nazeer2023soft, nazeer2023imitation}.
This imitation learning strategy (\autoref{fig:cover_pic}a), tailored for combinations of rigid and soft manipulators in series, extracts structured behaviours from minimal human demonstration.
The strategy enables generalization from a single demonstration to varying object and target configurations by explicitly exploiting the complementary roles of the two subsystems: repeatable motion generation by the rigid arm and physical, passive adaptation via the soft body's compliance (\autoref{fig:cover_pic}b).

This is thus our main contribution: by combining this learning framework with the proposed hybrid platform, we demonstrate that a wide range of manipulation skills can be acquired from a single human demonstration and executed robustly across task variations (\autoref{fig:cover_pic}d).
These include object stacking, non-prehensile manipulation, dynamic grasping, and grasping through narrow openings.
Across these tasks, the hybrid architecture naturally distributes functionality between its rigid and soft components.
Tasks that primarily demand accurate positioning and repeatability, such as stacking, are governed by the rigid arm and achieve performance comparable to that of a fully rigid manipulator.
In contrast, tasks involving geometric constraints or contact-rich interactions reveal the functional contribution of the soft arm, enabling manipulation strategies that cannot be realized with a rigid gripper alone.
\section*{Results}
\subsection*{The proposed system: combining rigid and soft robotics with imitation learning}
\begin{figure}[t]
    \centering
    \includegraphics[width=0.99\linewidth]{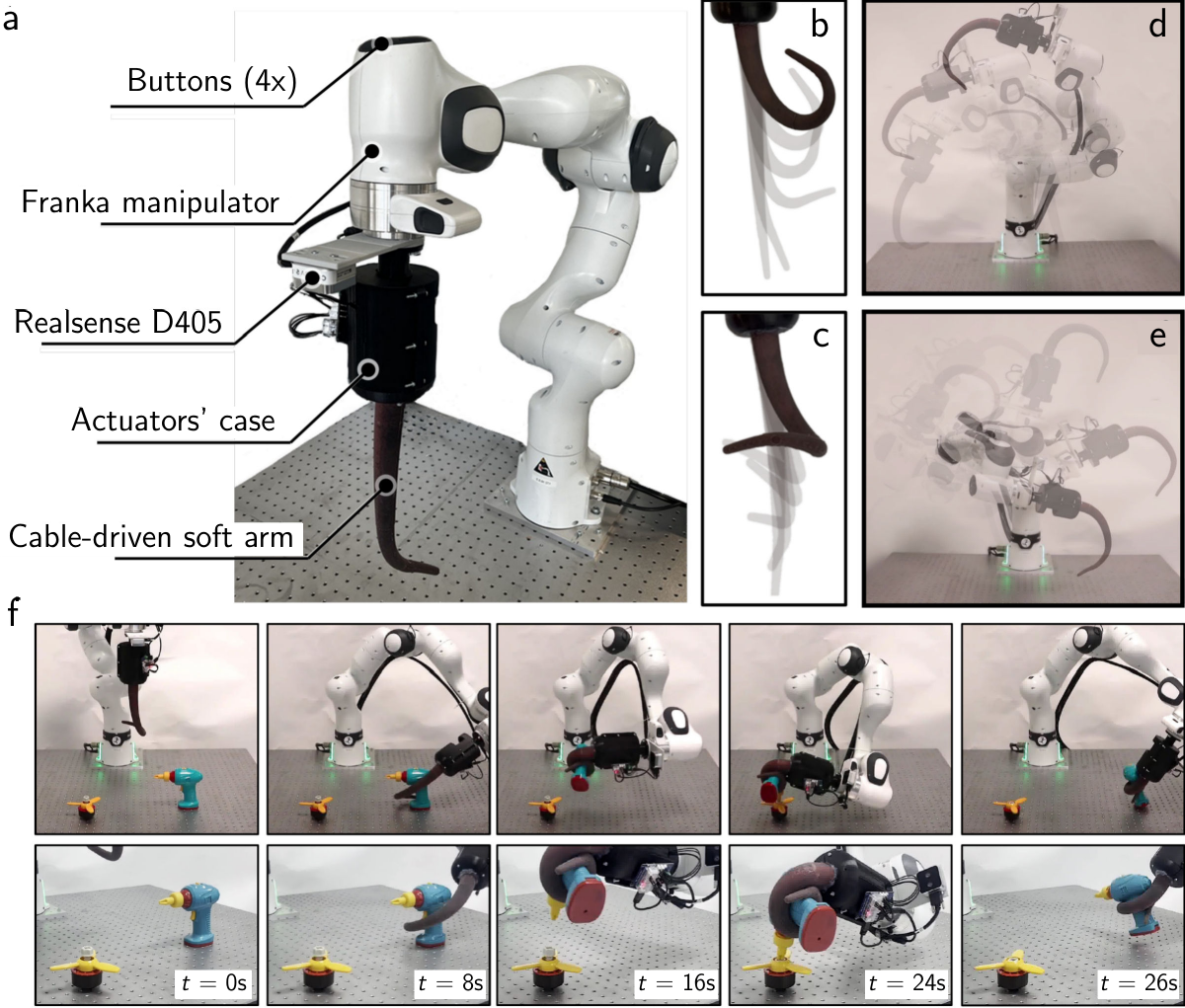}
    \caption{\textbf{Proposed soft–rigid hybrid platform.}
\textbf{a} The integrated system, composed of a 7-DoF rigid co-bot (Franka Panda), an RGB-D camera (RealSense D405) for keypoint localization, and a cable-driven soft arm mounted at the wrist. The soft arm is actuated via two tendons by motors enclosed in the actuator case. Four buttons on the robot are used to control and log tendon references during demonstrations. \textbf{b} Ventral bending of the soft arm when pulling one of the tendons. \textbf{c} Twisting of the soft arm when pulling the other tendon. Panels \textbf{d} and \textbf{e} show the platform executing compound motions where both rigid and soft components are actuated simultaneously. \textbf{f} A \emph{qualitative} example of a task we can do with our platform, drilling with a toy assembly.}
    \label{fig:setup}
\end{figure}

As discussed in the introduction, the proposed system integrates a rigid manipulator, a tendon-driven soft arm, and an imitation learning framework into a single control and execution pipeline, enabling learned behaviours to be distributed across mechanical and computational components.
The mechanical hardware, shown in \autoref{fig:setup}a, consists of a 7-DoF rigid arm (Franka Emika Panda) serially connected to a cable-driven soft arm with embedded tendons.
The soft arm draws inspiration from the morphology of the octopus.
It can bend and twist, as shown in \autoref{fig:setup}b and \autoref{fig:setup}c, and passively adapt to various contact situations, as shown in \autoref{fig:cover_pic}b.
The rigid arm ensures repeatable positioning and provides a reliable interface that enables kinesthetic demonstrations.
\autoref{fig:setup}d and \autoref{fig:setup}e show examples of coordinated motions of the whole system.
Additional technical details on the platform are provided in the Supplementary Materials.

The control layer is built on a custom imitation learning strategy that operates based on a single human demonstration and produces references for a low-level collocated control loop (see Supplementary Note 3).
Using spatial keypoints tracked via AprilTags and an RGB-D camera (RealSense D405), along with recorded end-effector poses and tendon states, the system infers a control policy that generalizes across object poses and task conditions (See Materials and Methods and Supplementary Note 4 for more details).
The rigid arm executes the retargeted motion, while the soft arm compensates for residual errors and environmental variation through its compliance.
Thus, the system is designed to leverage both rigid and soft components. Tasks such as stacking, pushing, or grasping in constrained spaces (see \autoref{fig:grasping_tasks} and \autoref{fig:non_grasping}) can be learned from a single demonstration and generalized without requiring retraining.
\autoref{fig:cover_pic}c and \autoref{fig:cover_pic}d illustrate qualitatively what we will detail quantitatively in the next subsections: by combining physical and computational intelligence, the system can generalize task execution at both the logical and physical levels. Similarly, \autoref{fig:setup}f illustrates a qualitative example of an assembly task, which requires precision during motions, compliance during insertion, and coordination between all the components involved.

We also characterize the platform’s robustness by evaluating its payload-handling capabilities and its tolerance to perceptual uncertainty. In a pick-and-place task, we test 19 combinations of object mass ($75$–$300$\,g) and diameter ($4.5$–$7.5$\,cm), observing that success rates decrease with increasing mass and with extreme diameters, see \autoref{fig:heatmap}. In a separate test, we assess robustness to perception noise by injecting perturbations of $5$–$15$\,cm in random directions into the perceived object position, and comnpare our system with three other alternatives, see \autoref{fig:noise_plot}.
\subsection*{Manipulation tasks}
\begin{figure}
    \centering
    \includegraphics[width=0.99\linewidth]{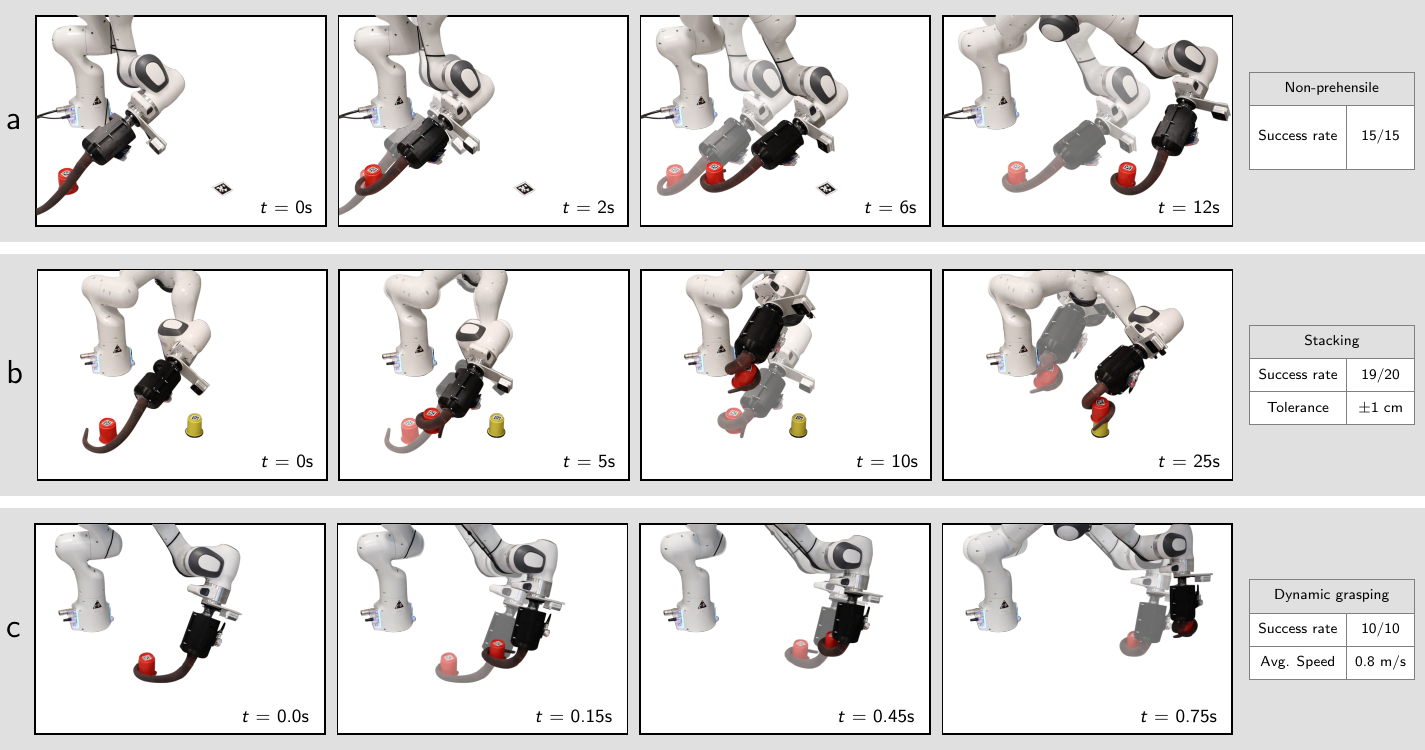}
    \caption{\textbf{A summary of our tested grasping tasks.} \textbf{a} One of the 10 tested configurations for the non-prehensile manipulation task. \textbf{b} One of the 20 tested configurations for the stacking task, and we also note the tolerance of $\pm1$ cm in this task due to the difference in radius of the two objects.  \textbf{c}~One of the 10 positions tested for dynamic grasping, and we also note the average speed of 0.8 m/s during the grasp. The success rate is also shown for all tasks in their respective panels.}
    \label{fig:grasping_tasks}
\end{figure}

We investigate two sets of manipulation skills.
The first set is focused on grasping: non-prehensile manipulation, stacking two cylindrical objects, and dynamic grasping, where we pick an object without pausing (see \autoref{fig:grasping_tasks}).
The generalizability of this set of skills is quantitatively tested by trying different configurations of the target objects, since we want to highlight precision and repeatability comparable to those of a conventional rigid robot.

The second set of skills focuses on tasks that are specifically enabled by our hybrid's fully soft arm: grasping an object and inspection through a narrow gap using a small camera mounted on the soft arm (see \autoref{fig:non_grasping}).
A typical rigid robot or typical soft gripper would not fit through the gap and/or be able to manipulate through it.
Due to the uniqueness of these tasks to our platform, we evaluate these tasks only qualitatively.

Finally, we also test unconventional grasping strategies in which the soft arm inserts itself into hollow objects and twists to manipulate them (see Supplementary Figure 5), and we qualitatively test the stacking of three objects (see Supplementary Note 6 and Supplementary Figure 3).
\subsubsection*{Non-prehensile manipulation}
\begin{figure}
    \centering
    \includegraphics[width=0.98\linewidth]{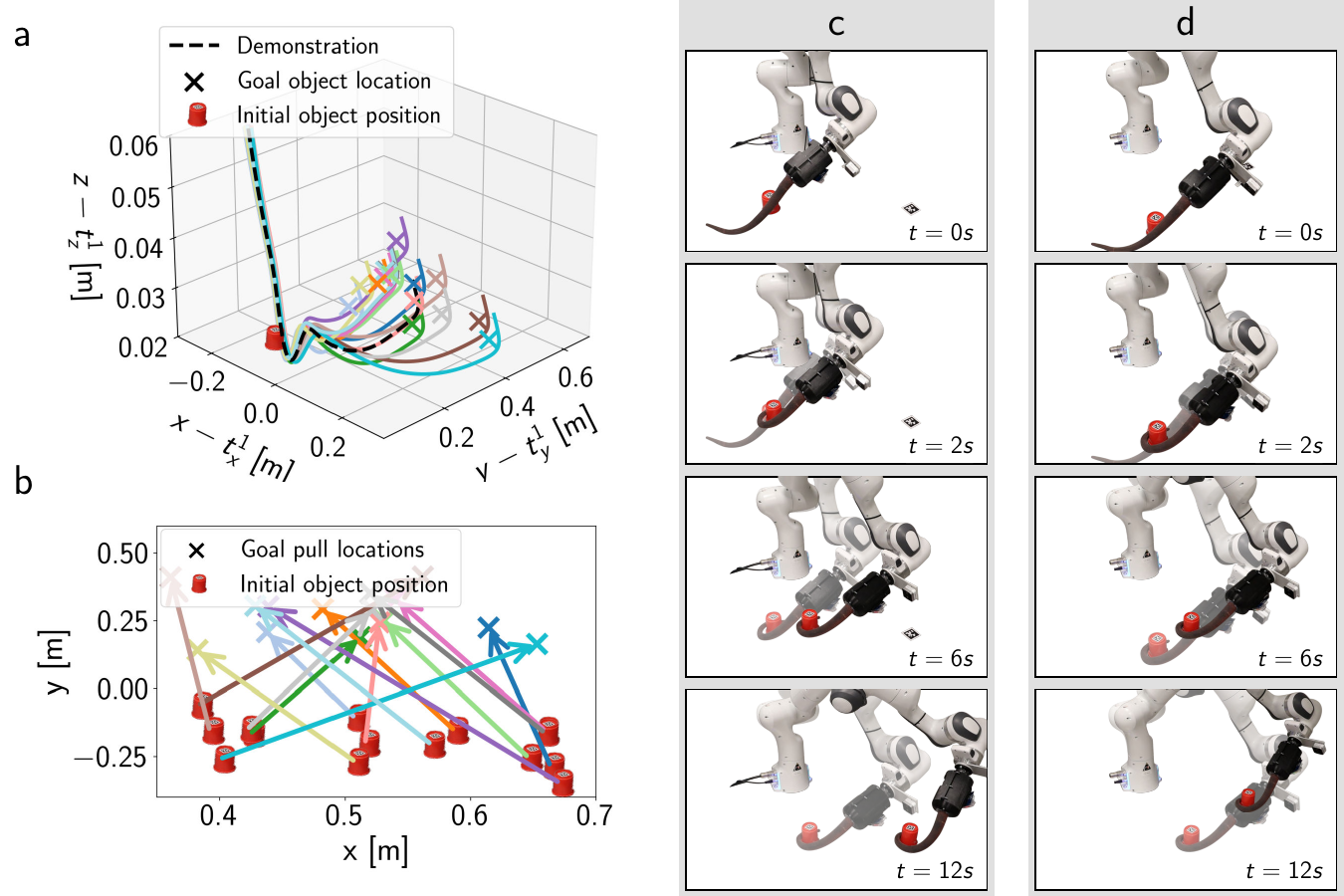}
    \caption{\textbf{Executions of the non-prehensile manipulation task.} \textbf{a}~Plot of the transported trajectories, where they have been spatially aligned by translating the picking locations to the origin for clarity. \textbf{b}~Initial and goal locations for the object from a top-view, with the arrows indicating paired locations. \textbf{c}~A sequence from an execution closely resembling the original demonstration, where the object is approached and dragged along a similar trajectory. \textbf{d}~A more demanding configuration, in which the robot must extend further to reach past the object before initiating the pull. In both instances, the manipulation is achieved without grasping, relying instead on compliant, non-prehensile contact. }
    \label{fig:pull_task}
\end{figure}
In this task we consider non-prehensile manipulation: a pulling task where the object is moved without being fully grasped, something which can be useful when dealing with objects that are too large or geometrically unsuitable for lifting, or to efficiently handle multiple items.
A representative execution is shown in \autoref{fig:grasping_tasks}a, where the hybrid can be seen approaching the object, hooking, and pulling it to the target goal.
While the soft arm is actuated to gently hook the object and steer it with precision into place, no lifting occurs; the object would be left behind if the robot were to move vertically.

We tested this left-to-right pulling behaviour across 15 randomly sampled combinations of start and goal positions, which are visualized in \autoref{fig:pull_task}b.
In all cases, the robot successfully completed the task.
\autoref{fig:pull_task}a also presents the trajectories in three dimensions.
Note that start and goal displacements vary non-rigidly across trials, rendering direct replay ineffective.

\subsubsection*{Stacking task}
\begin{figure}[h]
    \centering
    \includegraphics[width=0.98\linewidth]{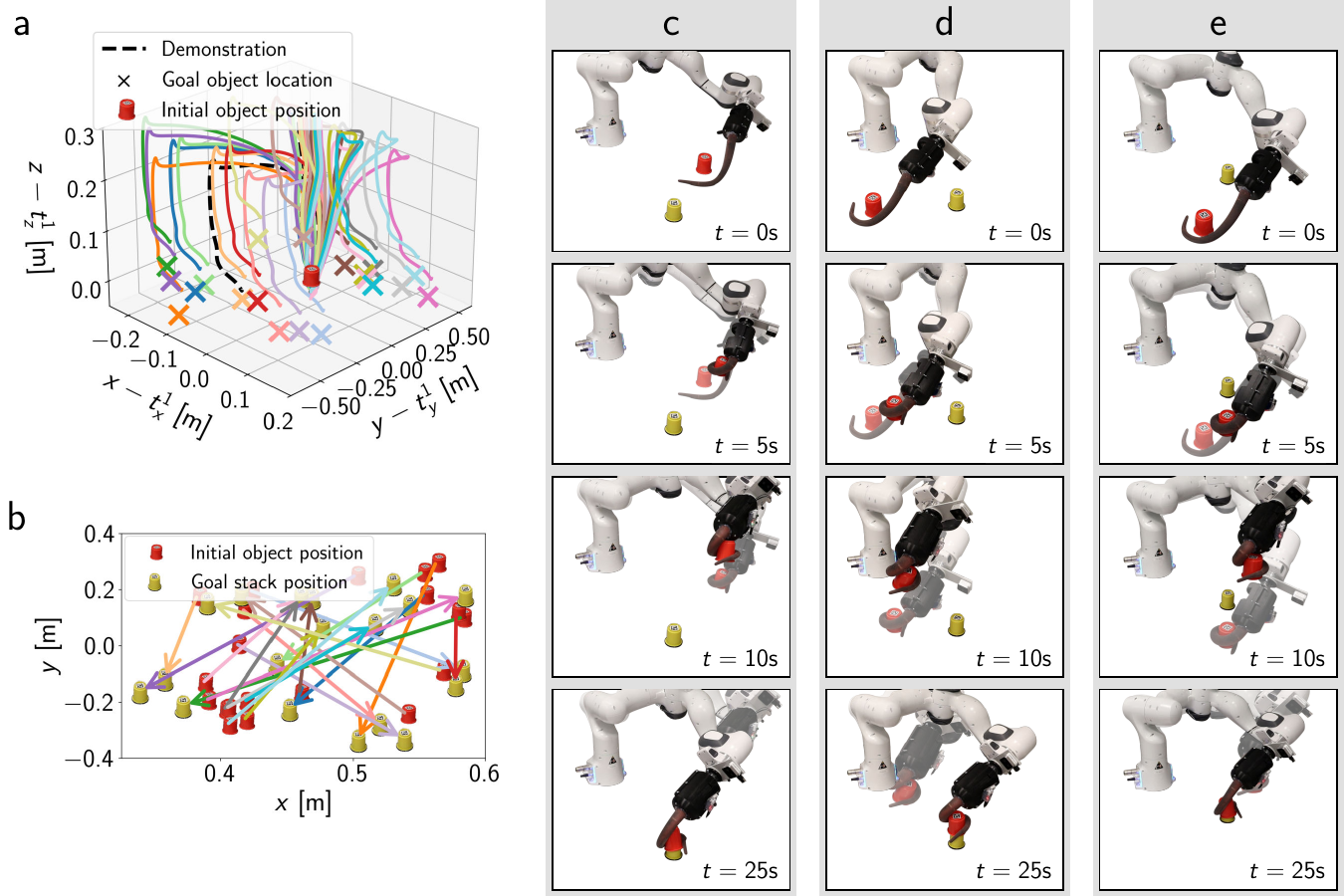}
    \caption{\textbf{Executions of the stacking task.} Panel (A) depicts the given demonstration, and transported trajectories in different colors, where these have all been spatially aligned to the picking locations for clarity. Panel (B) shows the initial object locations and the goal stacking locations from a top-view, with the colored arrows indicating the paired locations. Panels (C$_1$-C$_3$) visualize different initial configurations and their robot executions. Note how these different configurations showcase generalization, since the task is completed in configurations that are dissimlar to that of the demonstration.}
    \label{fig:stack_task}
\end{figure}
In the first task, we place two cylindrical cups of different radii in random locations in front of the robot, which must pick up the larger one and stack it on top of the smaller cup, as shown in \autoref{fig:grasping_tasks}b.
\autoref{fig:grasping_tasks}b also illustrates how, following a single demonstration, the robot can successfully reproduce the task when the configuration varies compared to the one in the demonstration --- here the robot must stack left-to-right, but the given kinesthetic demonstration involved stacking right-to-left. More execution sequences with varying configurations are shown in \autoref{fig:stack_task}c-e.

The principal generalization challenge stems from the \textit{non-rigidity} of the transformation: the two objects do not move as a single rigid body (i.e., preserving their relative posture), and thus the keypoint displacement is intrinsically nonlinear.
Handling such nonlinear mappings is a defining capability of the proposed architecture, as also shown by the varying configurations in the previous task. On the one hand, the learning module enables nonlinear re-targeting of the demonstrated motion, adapting it to altered spatial configurations. On the other hand, the physical intelligence embedded in the soft arm’s morphology compensates for discrepancies between the ideally optimal trajectory and the one actually produced by imitation.

We evaluated the task across 20 randomly sampled object configurations: 10 with the first object on the right and the second on the left, and 10 with the reverse arrangement.
All object configurations are shown in \autoref{fig:stack_task}b, and the 3D trajectories in \autoref{fig:stack_task}a.
Out of these, 19 trials were successful; the only failure occurred when the first cylinder was not stacked properly.
This demonstrates the consistent precision required for stacking, a task that is less forgiving than simply picking.
Softness naturally facilitates robust grasping by accommodating positional uncertainty, but it also enables compensation for misaligned object poses. The arm’s intrinsic compliance effectively serves as a physical counterpart to the commonly adopted strategy of Cartesian impedance control for such tasks \cite{ajoudani2017choosing,saveriano2023dynamic}.
\subsubsection*{Dynamic grasping}
\begin{figure}[h]
    \centering
    \includegraphics[width=0.98\linewidth]{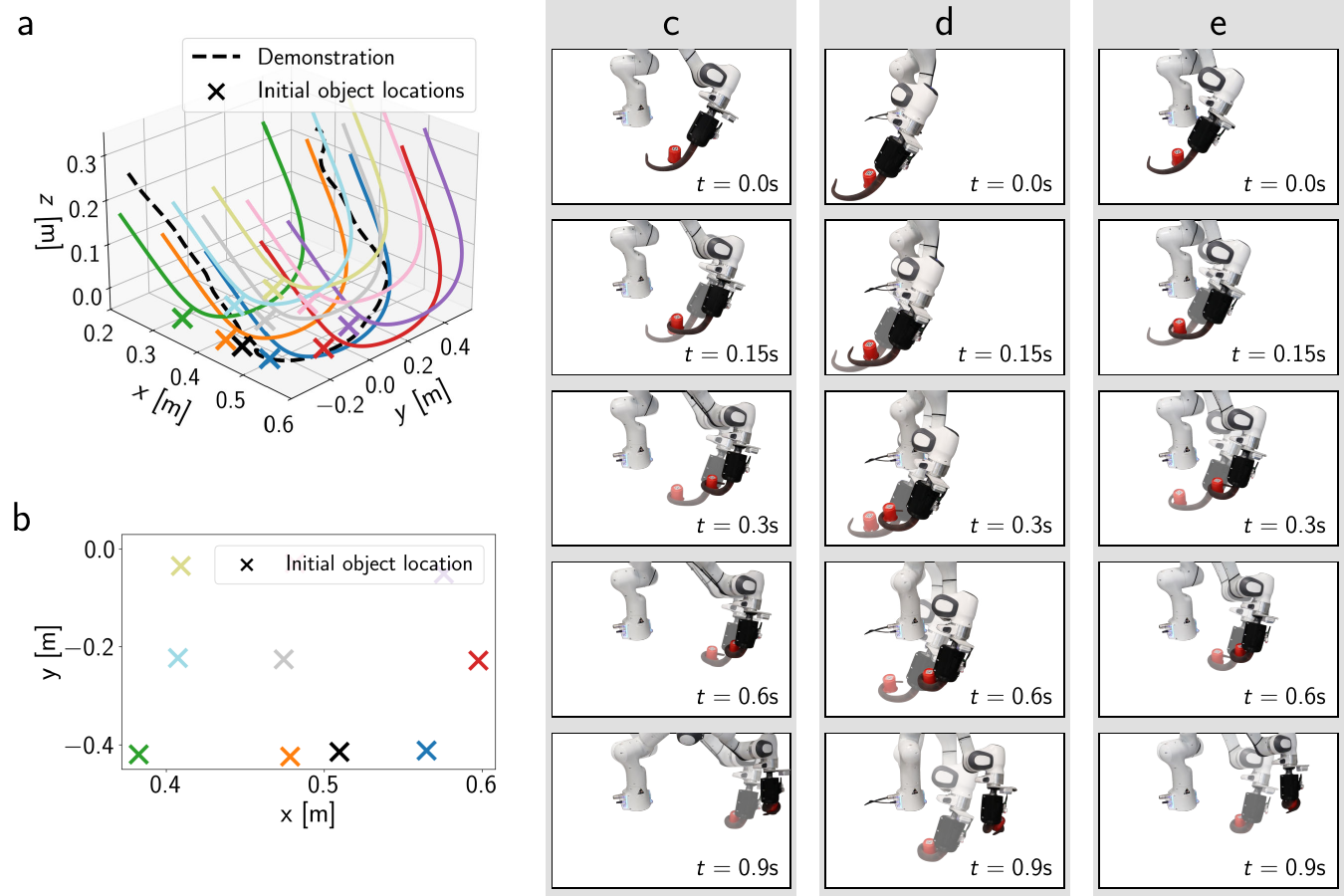}
    \caption{\textbf{Executions of the dynamic grasping task.} \textbf{a}~Demonstration and transported trajectories for each new configuration. \textbf{b}~Initial object locations from a top-view. \textbf{c, d, e} Several executions of this task, where the stills show initial contact, soft arm wrapping, and succesful picking. This task highlights the coordination between the fast motions that can be achieved via the rigid robot part and the physical adaptability of the soft arm.}
    \label{fig:dynamic_task}
\end{figure}
The goal in this task was to pick an object with non-zero velocity.
We tested this skill for 10 additional initial positions of the object, arranged in a roughly grid-like distribution as shown in \autoref{fig:dynamic_task}, where the robot succeeded every time.

While not difficult from a generalization point of view, this task mainly showcases the synergy between the rigid and the soft parts.
The rigid provides the required precision and speed for picking at non-zero velocities, while the compliance of the soft arm allows for a collision without issues.
The imitation learning component also allows us to implicitly program the coordination required between the soft arm's actuation and the rigid arm.
\autoref{fig:grasping_tasks}c shows stills of the robot performing this task, showing the initial collision, followed by the soft arm actuating and wrapping, and finally the object being picked.
\begin{figure}
    \centering
    \includegraphics[width=0.99\linewidth]{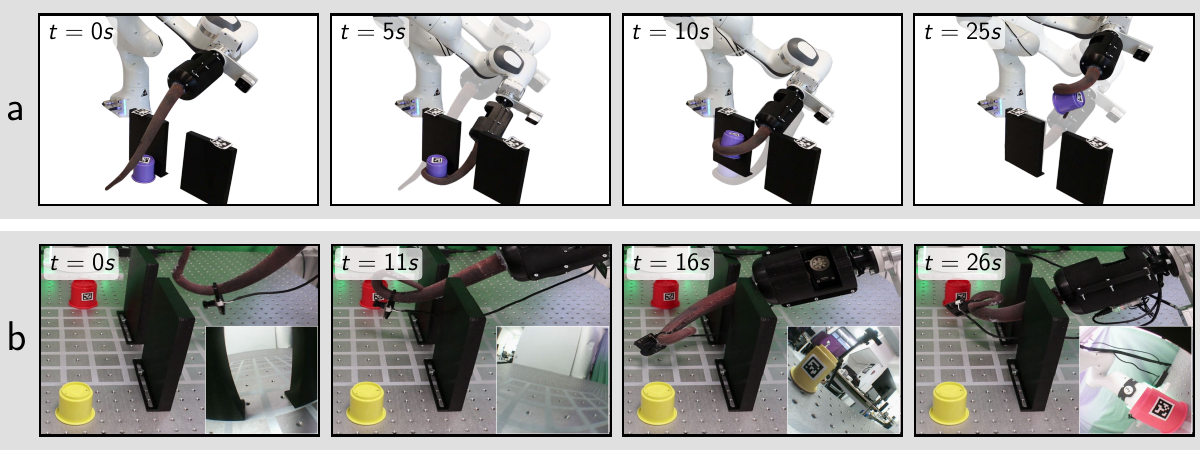}
    \caption{\textbf{A summary of our soft arm-specific tasks.} \textbf{a}~The hybrid platform manipulating an object through a narrow gap. \textbf{b}~The hybrid platform combining the movement of the rigid and soft parts to search for an object through a narrow gap using a small camera mounted on the soft arm; the view of this camera is shown in the bottom right of each frame.}
    \label{fig:non_grasping}
\end{figure}
\subsubsection*{Manipulation through narrow openings}
Apart from more conventional manipulation tasks, we wanted to highlight the cases where a conventional rigid robot would probably struggle with.
This task shows how we are able to slip the soft arm through a narrow opening to manipulate an object that is behind the obstacles creating the opening, see \autoref{fig:non_grasping}a.
Moreover, the soft arm allows us to perform this task with both smaller and larger objects using the same policy, as it naturally conforms to different sizes, see Supplementary Figure 4B-C.

We also test generalizability by changing the position and orientation of the walls.
As shown in the Supplementary Figure 4D, the task can still be completed, which also showcases the capability of the transportation algorithm to generalize not only the position but also the orientation of the end effector.
\subsubsection*{Inspection through narrow openings}
We also wanted to highlight the possibility for tasks soft-specific tasks that don't immediately involve grasping or manipulation, such as inspection.
In this qualitative example the hybrid can scan and inspect an otherwise unreachable area through a narrow gap by using a small camera mounted on the soft arm, as shown in \autoref{fig:non_grasping}b.
The hybrid can be used to move the camera around and look for the desired object, for example the red or yellow cylinders as shown in the mounted camera view on the bottom right.
Importantly, the camera's movement is enabled by both the rigid robot control and soft arm control.
Note that in this case this front-left-right scanning movement was also given through a demonstration.

\subsection*{Grasping under noisy observations}
We assess grasping performance under increasing levels of noise in the estimated object position.
We compare the soft arm to a rigid Franka gripper, a soft hand, and a soft gripper, revealing successful grasping at substantially higher noise levels.

\begin{figure}
    \centering
    \includegraphics[width=0.98\linewidth]{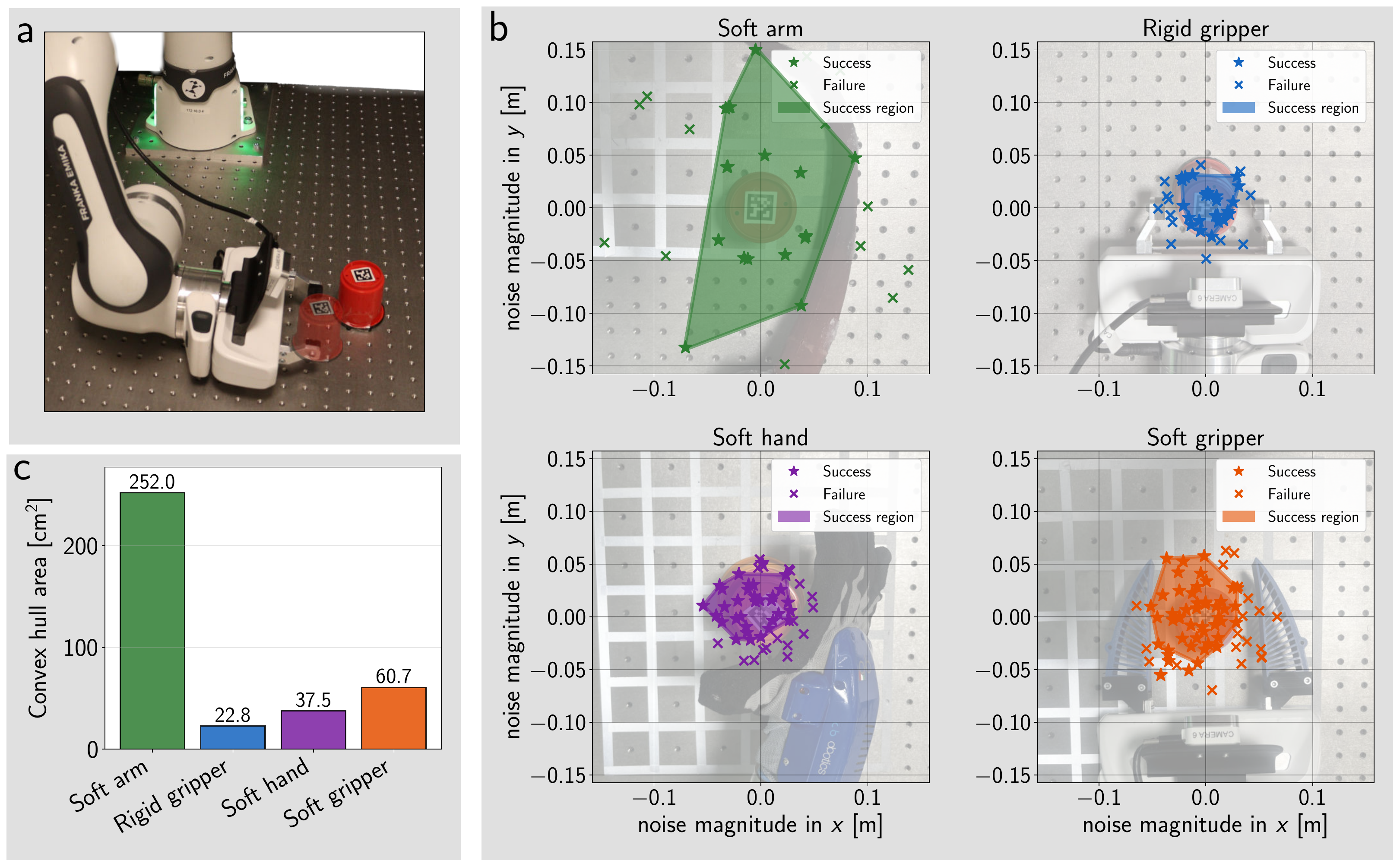}
    \caption{\textbf{Comparison between our hybrid platform, a rigid gripper, a soft hand, and a soft gripper when grasping with noisy observations}. \textbf{a}~An example of the rigid gripper tests, with the higher transparency cylinder showing where the robot believes the object is. \textbf{b}~Plotted noise vectors that were tested for each of these cases, showing which were successes or failures. A convex hull is also drawn using the successes for each case to approximate the success regions. \textbf{c}~The estimated areas for each of these cases shown in a bar plot; note how the hybrid has the largest area.}
    \label{fig:noise_plot}
\end{figure}

A noise vector of varying magnitudes and varying directions (as seen from above, i.e., on the $x$--$y$ plane) is added to the perceived object location when performing a grasping task.
The results are shown in \autoref{fig:noise_plot}, with the origin representing the real object location and the markers indicating where the object was perceived after noise and whether the grasp was successful or not.
For each case we draw convex hulls that are as large as possible while excluding any failures, which are meant to approximate the continuous noise region where objects can still be grasped.

As \autoref{fig:noise_plot}b shows, the region for the soft arm is much larger than the rest.
Note that while the comparatively larger size of the soft arm helps in having a larger success region, when fully wrapped the soft arm has a similar form factor to the compared alternatives.
More importantly, due to its full softness and the wrapping strategy (see e.g. \autoref{fig:grasping_tasks}), the exact position of the object is much less important compared to the alternatives, since the object can be pulled in from many different poses.
We also quantify the success region areas and show these in a bar plot in \autoref{fig:noise_plot}c, which shows the success region of the hybrid is more than 4 times larger than the next best alternative.

\subsection*{Payload characterization}
\begin{figure}[th!]
    \centering
    \includegraphics[width=0.98\linewidth]{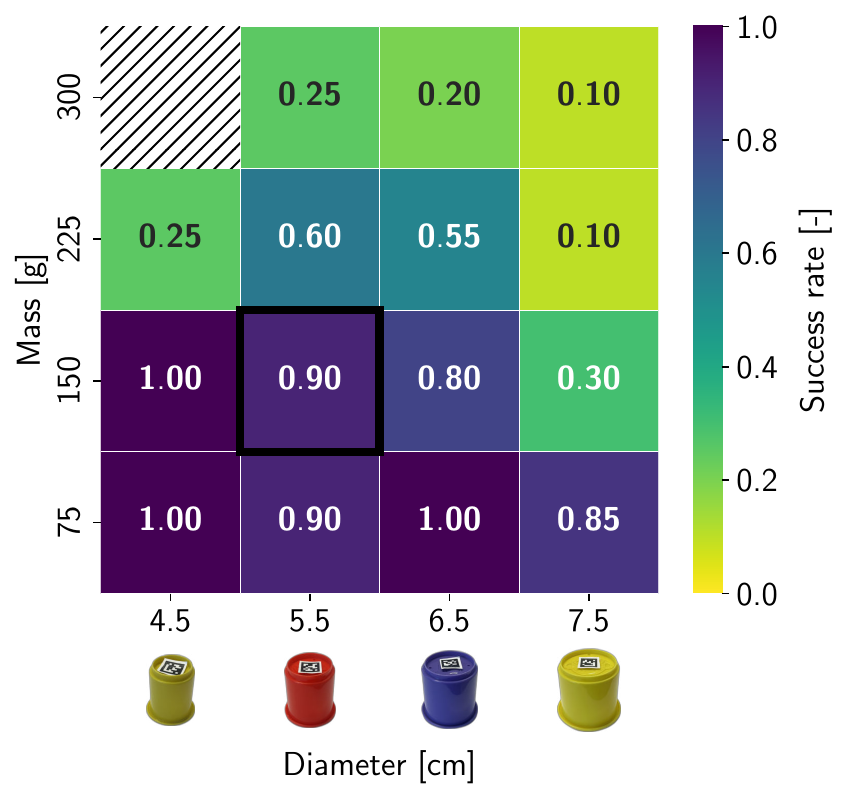}
    \caption{\textbf{Success rates for pick-and-place across object size and weight.}  Heatmap showing success rates over 10 trials for combinations of object diameter and mass. Lighter and smaller objects are handled reliably, while performance degrades as mass and size increase.}
    \label{fig:heatmap}
\end{figure}
We also test the platform's payload capabilities by investigating how varying object weight and size impacts the platform's grasping success.
We test this through a pick-and-place task, where the object starts at a random location on the left side of the robot and must be moved to a random location on the right side of the robot.
The task is repeated for different combinations of cylinder weight and diameter, and it is repeated 10 times per combination.
If the full task was completed, a score of 1 was given, if the object was picked but dropped before reaching the goal, a score of 0.5, and finally if the object was not picked up at all, a score of 0 was given.

We calculate a success rate for each combination of mass and diameter, and show these on a heatmap, see \autoref{fig:heatmap}.
The expectation was that the success rate would decrease with increasing weight, which is indeed shown in \autoref{fig:heatmap}.
The task also becomes harder for both too small or too large objects.
Since we are manipulating the object by wrapping around it, for too large objects the arm cannot completely wrap around and tighten around the object.
On the other hand, the soft arm has a limited radius it can curl to, meaning that objects that are smaller than this radius cannot be tightly grasped either.
\section*{Discussion}
The shown experiments provide evidence that combining soft and rigid components enables manipulation skills that can match the precision of a typical rigid robot, while unlocking new skills which are virtually impossible for a rigid robot or a typical soft gripper.
We achieve both precision and adaptability, spanning from conventional tasks, such as stacking, to soft-specific behaviours such as reaching through gaps.
The coordinated motions necessary to perform these tasks are possible thanks to a structure-preserving algorithm that generalizes a single demonstration by decoupling task retargeting, handled by the rigid arm, from physical passive adaptation, handled by the soft arm.
Moreover, we show how the synergy between the soft and rigid components is beneficial for tasks involving collision, like non-zero velocity picking, or to allow for compliance when stacking.
We also quantitatively show the benefit of picking strategies only available to the soft arm, such as wrapping, when there is uncertainty in perception.

Beyond performance, this work promotes a broader view of learning in robotics, where morphology is an active component of generalization.
The physical intelligence in the soft arm reduces the burden on computational learning by allowing meaningful behaviour to emerge from the interplay between control and structure.
This perspective opens new design opportunities: rather than relying solely on complex models or large datasets \cite{sunderhauf2018limits}, intelligence can be distributed across hardware and algorithms.
The result is a robust system that can consistently learn and perform tasks with an order of magnitude less data than state-of-the-art imitation policies \cite{chi2023diffusion,octo_2024,o2024open}.
Along this line, we envision future systems in which morphology, control, and learning are co-designed to achieve generalizable, data-efficient behavior~\cite{chen2020design,du2021diffpd,stolzle2025soft}.

While our approach emphasizes minimal data and physical generalization, some limitations and assumptions remain.
For example, keypoint-based parameterization provides a compact interface; however, pre-selecting keypoints and assuming that correspondences are known limits generalization.
Integrating pretrained visual-language-action models could help automate this process~\cite{brohan2023rt,kim2024openvla}.
Additionally, although our algorithm is highly efficient, incorporating multiple demonstrations could further enhance robustness and enable automatic inference of task structure.

Finally, from a mechatronic perspective, we foresee growing interest in unconventional robotic systems. We believe that more hybrid architectures, blurring the boundary between soft and rigid \cite{huang2025grasping, wang2025spirobs, koe2025learning, peng2025dexterous} and deliberative planning and embodied response \cite{chen2021co,drotman2021electronics,bhatt2022surprisingly,van2024bio,milana2025physical}, will be introduced.
Their capabilities will be leveraged by adaptive autonomy schemes that combine a blend of imitation learning, model-based control, and reinforcement learning.
These advancements will require innovations beyond the present work, like incorporating sensing into the soft arm, which would enable reactive feedback during contact-rich interactions.
Through this reactiveness, tactile sensing would enable policies that can grasp objects with different softness and fragilities.

Overall, this work demonstrates how combining soft and rigid components can provide both the precision of rigid robots and the adaptability of soft systems, enabling a broader range of manipulation behaviours.
By treating morphology as an active component of generalization, our approach illustrates how intelligence can be distributed between physical structure and algorithms.
Limitations remain in having to manually select keypoints and providing correspondences between source and target keypoints, and future work could explore adding tactile sensing to the soft body to use this as a feedback mechanism.
\section*{Methods}
\label{sec:method}
\subsection*{Hybrid robotic platform}
The mechatronic component of the proposed robotic system is a soft-rigid hybrid manipulator, comprising a rigid manipulator and a tendon-driven soft arm (\autoref{fig:setup}a). The rigid arm is a 7-DoF Franka arm that provides repeatable positioning and enables compliant kinesthetic teaching. To that, a cable-driven soft continuum soft arm is serially connected, inspired by the morphology of the octopus. The soft arm is molded from Ecoflex 00-50 and features a tapered geometry (35\,mm to 7\,mm over 380\,mm) to enhance its compliance and dexterity. The level of compliance in the soft structure is defined to ensure global asymptotic stability of the low-level closed loop while maintaining a high level of physical adaptability. Two internal tendons are actuated via Dynamixel MX-28 servos housed in a compact unit mounted at the robot's wrist. Actuating the two tendons generates the motions in \autoref{fig:setup}b and in \autoref{fig:setup}c. An Intel RealSense D405 RGB-D camera is positioned near the end effector and used to localize AprilTags on task-relevant objects, enabling visual generalization across task instances. Additional technical details and the fabrication description are given in Supplementary Note 1 and 2, respectively.
\subsection*{Experimental setups and protocols}
Each skill is learned from a single kinesthetic demonstration: the rigid arm is set to impedance control mode, allowing the user to manually guide its end-effector trajectory. Simultaneously, buttons mounted on the arm trigger discrete tendon actuation commands, which are executed by the servos and logged in sync with robot poses and camera-tracked keypoints. All demonstrations are recorded at 50\,Hz. Keypoints are defined by placing AprilTags on task-relevant objects and registering them via the D405 camera.  At execution time, the learned policy outputs both end-effector poses and tendon reference signals, which are issued through the Franka interface and a custom microcontroller, respectively. Tasks are executed with visual feedback only at initialization. For each skill, we evaluate generalization across unseen object poses and/or scene geometries, as detailed in the Supplementary Information.
\subsection*{Skill Representation and Learning for Rigid-Soft Hybrid}
\label{subsec:transportation}
The goal of this component is to provide a computational layer capable of learning, based on minimal data --- namely, a single demonstration --- how to autonomously execute task variations. It is responsible for generalization primarily at the logical level, while low-level adaptation and robustness are delegated to the soft robot’s (passive) embodied intelligence. We propose a learning strategy that builds upon our previous work~\cite{franzese2024generalization} and extends it to the context of soft robotics. This way, we enable the generalization of the recorded task into a skill, i.e., a policy capable of executing tasks within novel situations of the same class, such as varying object positions.

A core challenge lies in representing the internal state of the soft robot within an imitation learning framework. To address this, we adopt a rigid-robot-centric abstraction of the soft robot’s behavior during execution. This is made viable by assuming that the soft arm exhibits sufficient stiffness to guarantee a unique static equilibrium for each combination of base pose and tendon actuator configuration~\cite{della2025pushing}. The soft arm’s material properties were selected specifically to ensure this condition holds. Further details on this point, as well as on the low-level control of the soft arm, are provided in Supplementary Note 3. Under this assumption, the execution of a task is described as a time-resolved sequence of two directly actuated, repeatable quantities: the pose of the soft arm’s base in \(SE(3)\) and the tendon servo configuration in \(\mathbb{R}^{2+}\).

We refer to the structured representation of the task extracted from the single available demonstration as the set of \emph{policy labels}. Formally, the set of \(M\) Cartesian position labels is denoted by \({\mathcal{X}} = \{{x}_{1}, {x}_{2}, \ldots, {x}_{M} \}\), where each \({x}_i \in \mathbb{R}^3\). The corresponding set of orientation labels is \({\mathcal{R}} = \{{R}_{1}, {R}_{2}, \ldots, {R}_{M} \}\), with each \({R}_i \in SO(3)\). The set of servo configurations is denoted by \({\mathcal{L}} = \{{l}_{1}, {l}_{2}, \ldots, {l}_{M} \}\), where each \({l}_i \in \mathbb{R}^{2+}\).

We consider task classes that can be parameterized by a set of \(N\) Cartesian keypoints, which --- without loss of generality --- we take to be positions (though the method extends to full poses). For instance, in the stacking task, the keypoints are the object positions; in the grasping-behind-the-wall task, they include the positions of both the wall and the object. Varying these keypoints yields different instances within the same task class.
Let \(\tilde{\mathcal{S}} = \{\tilde{s}_{1}, \tilde{s}_{2}, \ldots, \tilde{s}_{N} \}\), with \(\tilde{s}_i \in \mathbb{R}^3\), denote the source keypoints corresponding to the demonstration. A new task instance is defined by a target keypoint set \(\tilde{\mathcal{T}} = \{\tilde{t}_{1}, \tilde{t}_{2}, \ldots, \tilde{t}_{N} \}\), with \(\tilde{t}_i \in \mathbb{R}^3\). Source and target keypoints are assumed to be registered—that is, \(\tilde{s}_i\) and \(\tilde{t}_i\) correspond to the same semantic element across instances.
A skill is thus a function that, given a new task specification \(\tilde{\mathcal{T}}\), generates a corresponding sequence of robot states. Our goal is to learn this function from a single demonstration, represented by the tuple \((\tilde{\mathcal{S}}, {\mathcal{X}}, {\mathcal{R}}, {\mathcal{L}})\).
\begin{figure}[t]
    \centering
    \includegraphics[width=0.75\linewidth]{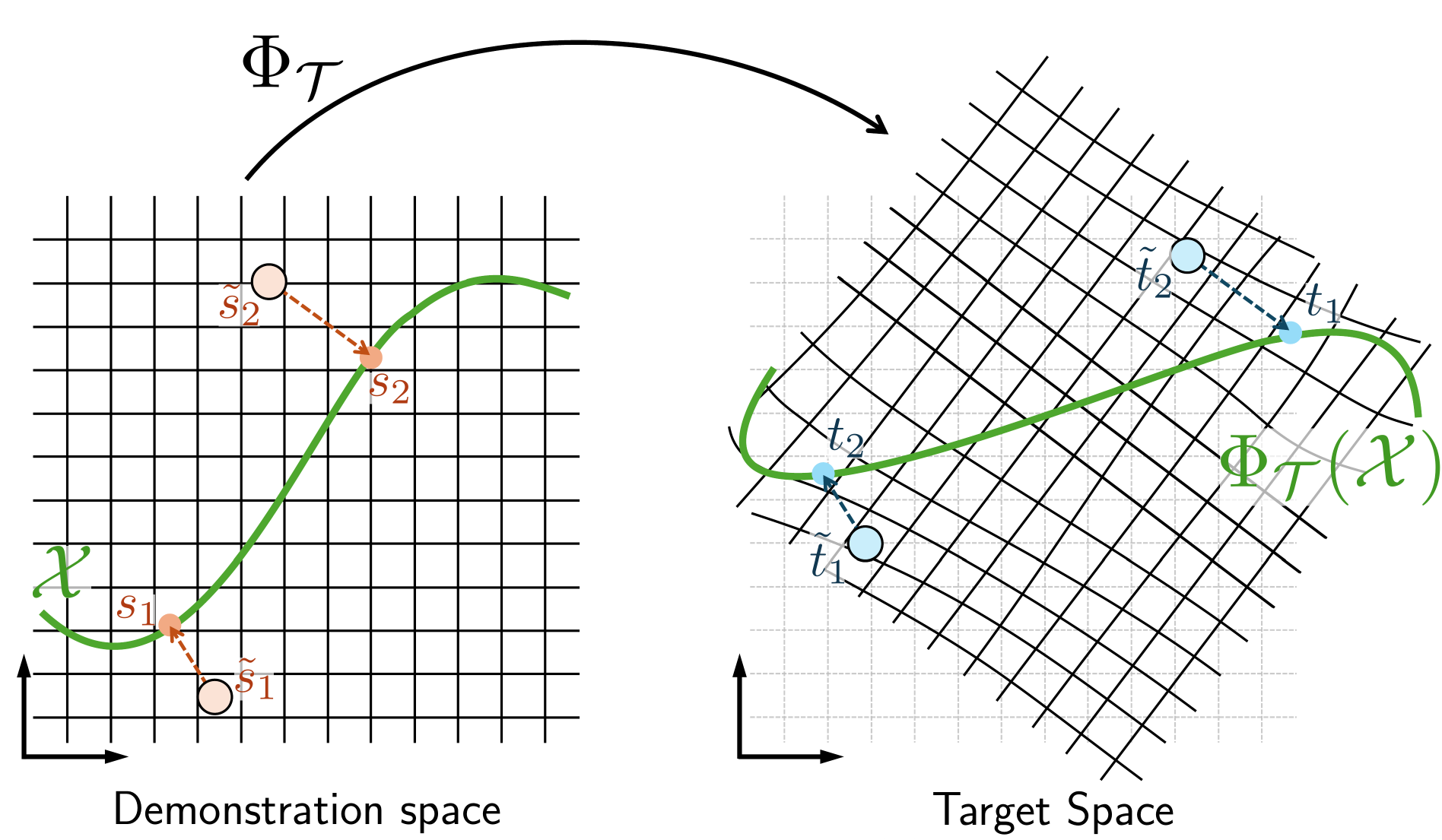}
    \caption{\textbf{Pictorial representation of the proposed algorithm.}
On the left we show the demonstration trajectory \(\mathcal{X}\), the source task-level keypoints \(\tilde{\mathcal{S}} = (\tilde{s}_1,\tilde{s}_2)\), and how they are re-targeted to robot-centric keypoints \(\mathcal{S} = ({s}_1,{s}_2)\). A new task instance is defined by \(\tilde{\mathcal{T}} = (\tilde{t}_1,\tilde{t}_2)\), which yields target robot-level keypoints \(\mathcal{T} = ({t}_1,{t}_2)\), shown on the right target space. A learned deformation map \(\bm{\Phi}_{\mathcal{T}}\) transports the original trajectory $\mathcal{X}$ to the generalized one in target space $\bm{\Phi}_{\mathcal{T}}(\mathcal{X})$  while preserving its structure.
}
    \label{fig:pictorialIL}
\end{figure}
The proposed algorithm operates in two steps: (i) re-target the task-space keypoints $\tilde{\mathcal{S}}$ and $\tilde{\mathcal{T}}$ — which are inherently difficult to handle due to the imprecise nature of soft manipulation—to corresponding keypoints at the rigid-robot level ${\mathcal{S}}$ and ${\mathcal{T}}$; and (ii) learn a \({\mathcal{T}}\)-parameterized deformation of the rigid robot's operational space and use it to map the demonstration \(({\mathcal{S}}, {\mathcal{X}}, {\mathcal{R}}, {\mathcal{L}})\) to the new task, see \autoref{fig:pictorialIL} for a visualization of this process.
The task-level keypoints \(\tilde{\mathcal{S}}\) and \(\tilde{\mathcal{T}}\) relate directly to the soft arm’s motion, as it is the component that physically interacts with the environment and ultimately carries out the task. However, due to the soft arm’s compliance and limited repeatability, we cannot assume reliable re-execution of the demonstrated motion. This is precisely where the rigid part of the hybrid platform becomes essential.
Thus, the first step of the algorithm maps the task-level keypoints to robot-centric keypoints. For each \({s}_i\), we find its nearest neighbor on the demonstrated Cartesian trajectory \({\mathcal{X}}\):
\begin{equation}
    {{s}}_i = \arg\min_{{x} \in {\mathcal{X}}} \lVert {x} - \tilde{s}_i \rVert.
    \label{eq:keypoint_remap}
\end{equation}
We denote by \({{\mathcal{S}}} = \{ {{s}}_1, \ldots, {{s}}_N \}\) the resulting ordered set of robot-centric keypoints.
We now define a corresponding set \(\mathcal{T}\) for the target task. To do so, we assume that the rigid manipulator should maintain the same relative displacement to the keypoints as in the demonstration. Therefore, each target keypoint is defined as ${t}_i = \tilde{t}_i + ({s}_i - \tilde{s}_i)$.

Now that the task keypoints have been expressed in a space that can be controlled with precision, we proceed to learn the skill. This is achieved by introducing a deformation map \({\Phi}_\mathcal{T}\) such that \({\mathcal{T}} = {\Phi}_\mathcal{T}({\mathcal{S}})\). We solve this supervised learning problem by first computing a rigid body transformation that brings \({\mathcal{S}}\) as close as possible to \({\mathcal{T}}\), and then fitting the residual deformation using Gaussian Process Regression. Further details are provided in the Supplementary Materials.
Once the map \({\Phi}_\mathcal{T}\) is learned, we use it to \emph{transport} the solution from the demonstration to the new task instance as:
\begin{equation}
    {\hat{\mathcal{X}}}(p) = {\Phi}_\mathcal{T}({\mathcal{X}}(p)), \quad
    {\hat{\mathcal{R}}}(p) = {J}_\mathcal{T}({\mathcal{X}}(p)) {\mathcal{R}}(p), \quad
    {\hat{\mathcal{L}}}(p) = {\mathcal{L}}(p),
\end{equation}
where \({J}_\mathcal{T} = \partial {\Phi}_\mathcal{T}({X}) / \partial {x}\) encodes the local rotation induced by the deformation at each point in space, and $p$ being our progress (time) variable.
Note that the inputs to the soft arm --- encoded in \({\mathcal{L}}(p)\) --- remain unchanged since they are not transformed by the map; generalization occurs exclusively on the physical side through passive adaptation.

In summary, the algorithm warps the entire task space of the rigid manipulator so that, in the deformed space, the robot's posture relative to the task keypoints matches that of the demonstration. Task execution can then be interpreted as the rigid robot re-playing the stored original trajectory \(({\mathcal{X}}(p), {\mathcal{R}}(p), {\mathcal{L}}(p))\) in a virtual, unwarped space, which corresponds to the generalized
\(({\Phi}_\mathcal{T}({\mathcal{X}}(p)), {J}_\mathcal{T}({\mathcal{X}}(p)) {\mathcal{R}}(p), {\mathcal{L}}(p))\)
when observed in the warped, physical space. Finally, it is also worth noting that this strategy directly extends to faster execution via simple time warping: \(({\Phi}_\mathcal{T}({\mathcal{X}}(\alpha p)), {J}_\mathcal{T}({\mathcal{X}}(\alpha p)) {\mathcal{R}}(p), {\mathcal{L}}(\alpha p))\), where \(\alpha \in (0,\infty)\).

\section*{Data availability}
The robot trajectories collected from demonstrations and associated source keypoints, the target keypoints for each tasks' executions, noise experiment data, and heatmap data can all be found at the repository \url{https://github.com/MRamirez25/dynamixel_control}.

\section*{Code availability}
Installation and usage instructions for all the Python packages and code we used can be found at the repository \url{https://github.com/MRamirez25/hybrid_ws} \cite{hybrid_ws_2026}.

\bibliography{sn-bibliography}

\section*{Funding}
This work was supported by the European Union's ERC grant RIPLEy No. 101165078.

\section*{Author Contributions}
M.R.M. contributed to the methodology and software, performed the robotic experiments and data analysis, conducted the visualization, and wrote the original manuscript and subsequent revisions. E.S. contributed to the methodology and soft arm fabrication, contributed to the visualizations, and contributed to the original manuscript and its revision. G.F. contributed to the methodology and software development and contributed to the original manuscript. J.K. and B.M. contributed to the conceptualization and supervision of the study and reviewed the manuscript. C.D.S. contributed to the conceptualization and supervision of the study, contributed to the visualization, and wrote and reviewed the manuscript. All authors reviewed and approved the final manuscript.
\section*{Competing interests}
The authors declare no competing interests.

\end{document}